\documentclass[lettersize,journal]{IEEEtran}
\usepackage{amsmath,amsfonts}
\usepackage{algorithmic}
\usepackage{algorithm}
\usepackage{array}
\usepackage[caption=false,font=normalsize,labelfont=sf,textfont=sf]{subfig}
\usepackage{textcomp}
\usepackage{stfloats}
\usepackage{url}
\usepackage{verbatim}
\usepackage{graphicx}
\usepackage{cite}

\usepackage{amssymb}
\usepackage{multirow,booktabs}
\usepackage[hidelinks]{hyperref}
\usepackage{tabularx}
\usepackage{subcaption}
\begin{document}

\title{Bi‑Directional Feedback Fusion for Activity‑Aware Forecasting of Indoor CO$_2$ and PM$_{2.5}$}

\author{Harshala~Gammulle,~\IEEEmembership{Member,~IEEE,}
        Lidia~Morawska,~\IEEEmembership{}
        Sridha~Sridharan,~\IEEEmembership{Life Senior Member,~IEEE,}
        Clinton~Fookes,~\IEEEmembership{SeniorMember,~IEEE.}

        \IEEEcompsocitemizethanks{
    \IEEEcompsocthanksitem H. Gammulle, S. Sridharan, and C. Fookes are with the Signal Processing, Artificial Intelligence and Vision Technologies (SAIVT) Lab, Queensland University of Technology, Brisbane, Australia.\protect\\
    E-mail: pranali.gammule@qut.edu.au
    \IEEEcompsocthanksitem L. Morawska is with the International Laboratory for Air Quality and Health (ILAQH), Queensland University of Technology, Brisbane, Australia.
}
}

\markboth{Journal of \LaTeX\ Class Files,~Vol.~14, No.~8, August~2021}%
{Shell \MakeLowercase{\textit{et al.}}: A Sample Article Using IEEEtran.cls for IEEE Journals}


\maketitle

\begin{abstract}
Indoor air quality (IAQ) forecasting plays a critical role in safeguarding occupant health, ensuring thermal comfort, and supporting intelligent building control. However, predicting future concentrations of key pollutants such as carbon dioxide (CO$_2$) and fine particulate matter (PM$_{2.5}$) remains challenging due to the complex interplay between environmental factors and highly dynamic occupant behaviours. Traditional data‑driven models primarily rely on historical sensor trajectories and often fail to anticipate behaviour‑induced emission spikes or rapid concentration shifts. To address these limitations, we present a dual-stream bi-directional feedback fusion framework that jointly models indoor environmental evolution and action-derived embeddings representing human activities. The proposed architecture integrates a context-aware modulation mechanism that adaptively scales and shifts each stream based on a shared, evolving fusion state, enabling the model to selectively emphasise behavioural cues or long-term environmental trends. Furthermore, we introduce dual timescale temporal modules that independently capture gradual CO$_2$ accumulation patterns and short-term PM$_{2.5}$ fluctuations. A composite loss function combining weighted mean squared error, spike-aware penalties, and uncertainty regularisation facilitates robust learning under volatile indoor conditions. Extensive validation on real-world IAQ datasets demonstrates that our approach significantly outperforms state-of-the-art forecasting baselines while providing interpretable uncertainty estimates essential for practical deployment in smart buildings and health-aware monitoring systems.
\end{abstract}


\section{Introduction}

The World Health Organization (WHO) identifies ambient and household air pollution as major global environmental health risks, responsible for millions of premature deaths each year \cite{WHO_2022_airpollution, kuehn2014more, WHO_2022_household}. In this context, although household air pollution refers specifically to pollution from solid fuel combustion, indoor air quality (IAQ) represents a broader concept encompassing ventilation, indoor pollutant sources, and environmental conditions that influence health, comfort, and cognitive performance in homes, classrooms, and offices \cite{seppanen1999association, mendell2006risk}.

Outdoor pollution continuously infiltrates indoor spaces, and in poorly ventilated buildings, indoor pollutant concentrations can even surpass outdoor levels \cite{CSIRO_2024_IAQ, kajjoba2024impact} due to indoor sources. Among the key IAQ pollutants, carbon dioxide (CO$_2$) and fine particulate matter (PM$_{2.5}$) are widely monitored because of their strong associations with ventilation efficiency, occupant density, and human activities \cite{persily2017carbon, ramalho2015association, liu2022systematic}. However, forecasting future concentrations of these pollutants is inherently challenging due to the complex interplay between environmental factors and dynamic occupant behaviours.

Traditional IAQ forecasting models primarily rely on historical sensor readings (e.g., temperature, humidity, CO$_2$, PM$_{2.5}$) and employ statistical or deep learning approaches such as recurrent neural networks (RNNs) or convolutional architectures \cite{wei2019machine, bakht2022deep}. While these models effectively capture temporal patterns and lagged dependencies from environmental signals, they often overlook latent contextual and behavioural drivers of pollutant dynamics. For instance, occupant actions such as cooking, cleaning, window-opening, or abrupt occupancy changes can trigger rapid PM$_{2.5}$ spikes or CO$_2$ surges that are not predictable from historical sensor trajectories alone. Consequently, models trained solely on conventional sensor data may exhibit degraded generalisation when confronted with behaviour-driven emission events or shifts in occupant patterns.


To address this gap, we propose a novel \emph{bi-directional feedback fusion framework} that explicitly integrates environmental sensing streams with action-based embeddings representing occupant behaviours. Our approach formulates IAQ forecasting as a two-stream temporal modelling problem for predicting future pollutant concentrations (CO$_2$ and PM$_{2.5}$). One stream captures environmental evolution, while the other encodes human actions through language-derived embeddings. A shared fusion mechanism iteratively aligns these modalities via bi-directional feedback loops, enabling joint reasoning over environmental and behavioural dynamics.

Unlike conventional concatenation or attention-based fusion strategies, our architecture introduces a \emph{context-aware modulation} mechanism, wherein the global fusion state adaptively scales and shifts each stream based on evolving conditions. This design allows the model to amplify action cues during transient PM$_{2.5}$ events or prioritise long-term environmental context during stable occupancy periods. Furthermore, to capture dependencies across different time horizons, we employ \emph{dual timescale modules}: a long-term branch modelling gradual CO$_2$ trends and a short-term branch responding to rapid transient PM$_{2.5}$ changes.

Finally, we formulate a composite learning objective that combines weighted mean squared error with a spike-aware penalty and an uncertainty-aware term, ensuring balanced optimisation across pollutants with distinct temporal and statistical properties. The resulting model not only achieves higher forecasting accuracy but also provides interpretable uncertainty estimates, which are critical for practical deployment in smart building control and health monitoring systems.

In summary, this work makes the following key contributions:
\begin{itemize}
    \item We propose a dual-stream bi-directional feedback fusion framework for IAQ forecasting that jointly models indoor environmental signals and action embeddings representing occupant behaviours.
    \item We introduce a context-aware modulation mechanism that adaptively refines stream-specific representations based on the evolving global fusion state.
    \item We design dual timescale temporal modules to capture both long-term contextual trends and short-term action-induced variations in pollutant dynamics.
    \item We formulate a composite training objective that integrates spike-aware weighting and uncertainty-based regularisation, enhancing robustness and interpretability under volatile indoor conditions.
    \item Extensive experiments on real-world indoor datasets demonstrate significant improvements over state-of-the-art temporal forecasting baselines for both CO$_2$ and PM$_{2.5}$ predictions.
\end{itemize}

\section{Related Work}
Air pollution poses severe risks to human health, making effective mitigation strategies essential~\cite{WHO_2022_household,WHO_2022_airpollution}. Outdoor air pollution, driven by industrial emissions, vehicular exhaust, and urbanization, significantly influences indoor air quality (IAQ), especially since people spend approximately 90\% of their time indoors~\cite{morawska2022healthy}. In poorly ventilated environments, indoor emitted pollutant concentrations can even exceed outdoor levels~\cite{CSIRO_2024_IAQ, kajjoba2024impact}. IAQ is further affected by indoor sources and human activities; for instance, severe outdoor particle pollution elevates indoor PM$_{2.5}$ concentrations, while domestic cooking is a major contributor to particulate matter~\cite{liu2022systematic}.

Early IAQ forecasting approaches relied on physics-based and statistical models, such as mass-balance formulations and multilinear regression, which estimated pollutant accumulation based on ventilation rates, emission coefficients, and historical observations~\cite{Persily_2017_IndoorAir,Wei_2019_IndoorAir}. While interpretable, these models required manually specified parameters, limiting adaptability across diverse environments. Recent research has shifted toward machine learning (ML) and deep learning (DL) methods that learn nonlinear pollutant dynamics directly from sensor time series (e.g., temperature, humidity, CO$_2$, PM$_{2.5}$, NO$_2$). Classical ML algorithms such as Random Forests (RF), Support Vector Machines (SVM), and Gradient Boosting (GBM) improved predictive accuracy over empirical formulations~\cite{Wei_2019_IndoorAir}. Building on this, deep learning architectures such as Recurrent Neural Networks (RNNs), Long Short-Term Memory (LSTM) models, and Temporal Convolutional Networks (TCNs) captured long-term temporal dependencies in IAQ data~\cite{Bakht_2022_Toxics}. More recently, Transformer-based models~\cite{vaswani2017attention} have been applied for multi-horizon pollutant forecasting, leveraging self-attention mechanisms for temporal encoding~\cite{Zhang_2024_STE,laton2025artificial}. Despite these advances, most existing models rely solely on environmental sensor readings and overlook contextual drivers such as occupant activities, leaving substantial room for improvement in behaviour-aware IAQ forecasting frameworks~\cite{alsamrai2024systematic,saini2020indoor}.

Emerging efforts have begun to recognise occupant behaviour as a critical driver of pollutant dynamics. For example, Lin et al.~\cite{lin2017analyzing} demonstrated measurable associations between activities (e.g., cooking, window opening, occupancy patterns) and IAQ variations using smart-home sensor logs and chemical measurements. However, only a minority of IAQ forecasting models incorporate behavioural features such as occupancy counts~\cite{dai2024method} or appliance usage and ventilation events~\cite{kim2022prediction}. This gap underscores the need for forecasting architectures that treat occupant behaviour as a distinct input stream. In this work, we propose modelling action embeddings as a parallel stream to environmental sensor readings, enabling joint reasoning over behavioural context and ambient measurements.

Multimodal fusion has been widely explored across domains \cite{perez2018film,tsai2019multimodal,xu2024review}. Early fusion strategies relied on feature concatenation, which limits modality interaction. Conditioning mechanisms~\cite{perez2018film} and cross-modal attention~\cite{tsai2019multimodal} introduced dynamic space for emphasizing relevant cues. Transformer-based multimodal fusion has been reviewed in recent work~\cite{xu2024review}, highlighting attention-driven mechanisms for robust integration. More recent methods adopt bi-directional fusion, iteratively refining representations through reciprocal conditioning to improve robustness in sensor and vision-sensor fusion tasks, such as LiDAR-radar fusion~\cite{wang2023bi}. Inspired by these developments, we adopt a bi-directional feedback fusion architecture for IAQ forecasting, enabling mutual modulation between environmental and behavioral streams. Our design also incorporates shared-private frameworks~\cite{wang2025multi,van2023aspnet} but enhances them with iterative feedback loops for dynamic alignment between modalities.

\section{Method}
\subsection{Overview}

Our goal is to forecast indoor CO$_2$ and PM$_{2.5}$ levels by combining two complementary information streams:
\begin{itemize}
    \item {Environmental Stream}: Indoor environmental sensor readings which includes temperature, humidity, pollutant levels.
    \item {Action Stream}: Human activity embeddings derived from occupant behavior logs.
\end{itemize}

The proposed framework fuses these streams through a bi-directional feedback mechanism, then models temporal dependencies using dual timescale modules for long-term and short-term patterns. Finally, predictions are generated with uncertainty estimates for practical deployment.

\subsection{Inputs and Problem Setup}

For the two input streams, $X^{e}$ and $X^{a}$, we adopt a sliding-window approach with a lookback window of length $L$ (past measurements used for prediction) and a prediction window of length $P$ (future time steps to forecast). Given $L$ and $P$, the input can be defined as:

\begin{equation}
   X = [X^{e}, X^{a}] \in \mathbb{R}^{L \times D},
\end{equation}

where $X^{e} \in \mathbb{R}^{L \times d_{e}}$ and $X^{a} \in \mathbb{R}^{L \times d_{a}}$ represent the environmental and action streams, respectively. The output $Y \in \mathbb{R}^{P \times 2}$ provides the future CO$_2$ and PM$_{2.5}$ levels for $P$ time steps.

\subsection{Model Architecture}

We design the model architecture to capture stream-specific characteristics while maintaining a bi-directional, feedback-based fusion mechanism that enables complementary learning between the two streams. This is particularly important because fluctuations in one pollutant stream (e.g., rising $CO_{2}$ due to increased occupancy) can precede and inform the behaviour of the other (e.g., elevated $PM_{2.5}$ from subsequent cooking activities).

\begin{figure}[htbp]
        \centering
        	\includegraphics[width=1.0\linewidth]{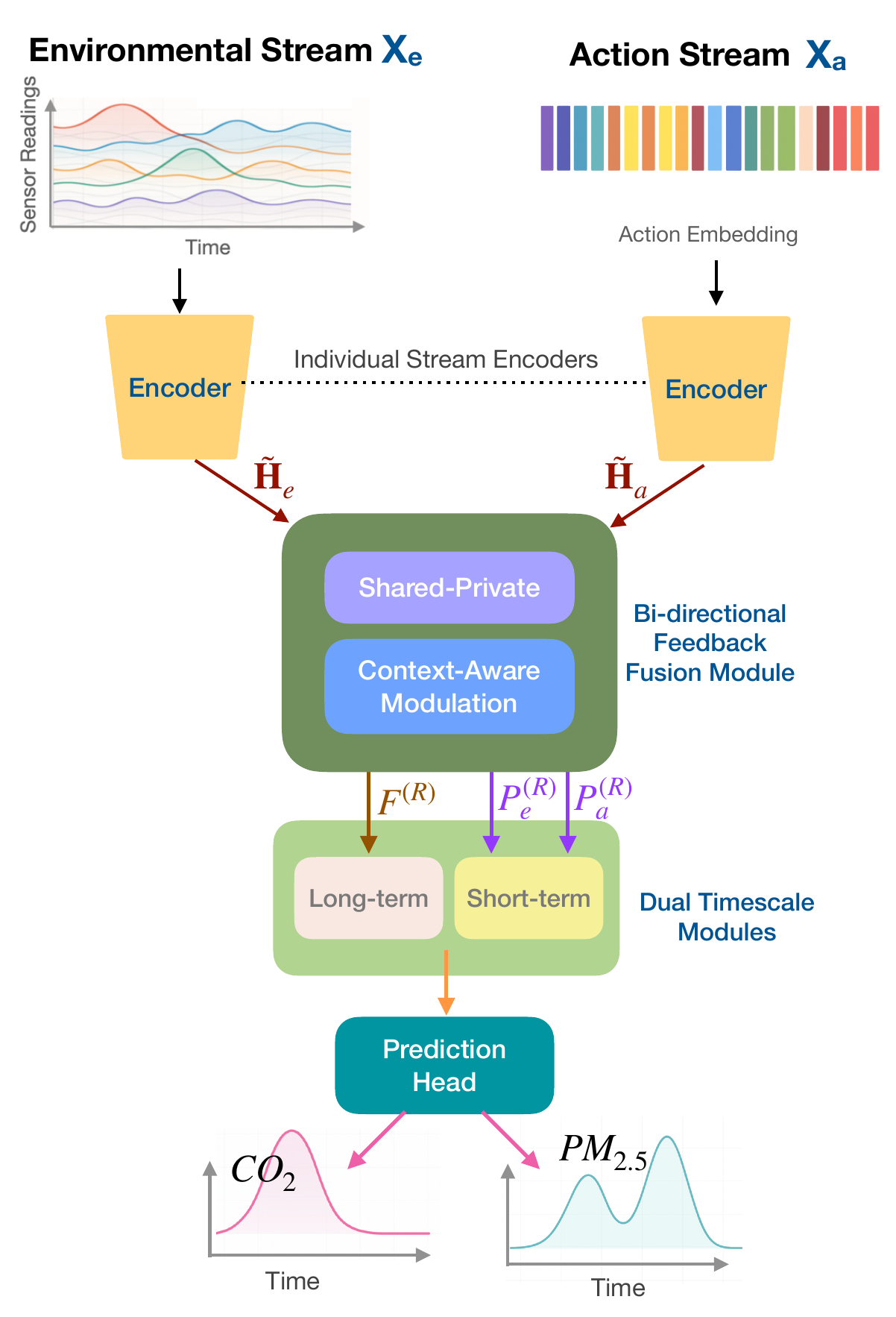}
	\caption{Overview of the proposed dual‑stream bi-directional feedback fusion framework for indoor air‑quality forecasting. The model processes environmental sensor data and action embeddings through separate stream encoders to capture environmental evolution and human activity patterns. A bi-directional feedback fusion module iteratively aligns and refines cross‑stream representations using shared‑private decomposition and context‑aware modulation. Dual‑timescale modules then model both long‑term CO$_2$ trends and short‑term PM$_{2.5}$ fluctuations before generating pollutant-specific predictions.}
	\label{fig:overview}
\end{figure}

The proposed architecture consists of five key components: Individual Stream Encoders, Shared and Private Representations, Bi-Directional Feedback Fusion, Dual-Timescale Temporal Modules, and the Prediction Head.

\subsubsection{Individual Stream Encoder}

Prior to fusion, it is crucial to model the inherent temporal dynamics within each stream independently. To achieve this, we employ an encoder that projects the original features into a lower-dimensional latent space while preserving semantic dependencies necessary for cross-stream interaction. In other words, each stream is encoded separately to capture its unique temporal characteristics: the environmental encoder learns environmental trends, whereas the action encoder learns behavioral patterns.

For a stream \( q \) (where \( q \) represents either the environmental (\( e \)) or action (\( a \)) stream), the input first passes through a linear projection layer, followed by a Stream-wise Temporal Encoder (STE), denoted as \( \mathcal{T}_{\text{STE}_q} \), and finally another linear projection for alignment. The STE is adopted from our prior work \cite{gammulle2025dual} due to its enhanced temporal modeling capability, which effectively captures stream-specific dynamics and complements the overall feature mapping for pollutant level forecasting. Formally, this process can be expressed as:

\begin{equation}
U_q = \mathrm{ReLU}\!\left( X_q W_p^{(q)} + b_p^{(q)} \right) \in \mathbb{R}^{L \times h},
\label{eq:proj_layer}
\end{equation}

\begin{equation}
H_q = \mathcal{T}_{STE_q}(U_q),
\quad H_q \in \mathbb{R}^{L \times h},
\label{eq:temporal_encoder}
\end{equation}

\begin{equation}
\tilde{H}_q = H_q W_o^{(q)} + b_o^{(q)} \in \mathbb{R}^{L \times h}.
\label{eq:output_proj}
\end{equation}

Here, \( W_p^{(q)} \) and \( W_o^{(q)} \) are the weight matrices for the input and output projections, \( b_p^{(q)} \) and \( b_o^{(q)} \) are bias terms, \( L \) denotes the lookback window length, \( d_{\text{in}} \) is the input dimensionality for each stream, and \( h \) represents the hidden feature dimension. 

For the action stream, which consists of short text-based activity descriptions, we first embed these labels using the \textit{all-MiniLM-L6-v2} model from the Sentence-Transformers library \cite{wang2020minilm}. This choice was motivated by its strong performance on semantic similarity tasks involving short phrases, which aligns with the nature of our activity labels. MiniLM provides compact 384-dimensional embeddings that capture contextual semantics efficiently, ensuring robust representation without introducing significant computational overhead. The resulting embeddings are subsequently processed by the stream encoder to capture temporal semantic action cues.

\subsubsection{Shared and Private Representations}

Once the stream-specific temporal information is learned through the individual stream encoders ($U_{e}$ and $U_{a}$), each stream is linearly transformed into two parallel subspaces. For each stream, the model learns two simultaneous feature projections: one emphasizing shared patterns and the other emphasizing private (stream-specific) cues. The shared representation captures common patterns (e.g., occupancy effects), while the private representation captures stream-specific cues (e.g., cooking spikes for PM$_{2.5}$). We believe this separation prevents one modality from dominating the other.

\begin{align}
S_e &= \tilde{H}_e\, W^{(e)}_{S} + b^{(e)}_{S}, &
S_a &= \tilde{H}_a\, W^{(a)}_{S} + b^{(a)}_{S}, \\
P_e &= \tilde{H}_e\, W^{(e)}_{P} + b^{(e)}_{P}, &
P_a &= \tilde{H}_a\, W^{(a)}_{P} + b^{(a)}_{P}.
\end{align}

Here, $S_e$ and $S_a$ represent the shared candidates from each stream (environmental and action), which capture common latent attributes that jointly explain the behaviors of both pollutants, while $P_e$ and $P_a$ represent the private (i.e., stream-specific) candidates, which capture unique semantics within each stream. Furthermore, $W^{(\cdot)}_{S}$ and $W^{(\cdot)}_{P}$ are the learnable matrices that determine how much of the temporal feature space of each stream contributes to the shared versus private components.

To obtain the global fused representation (denoted as $F^{(0)}$), the model combines the two shared candidates as:

\begin{equation}
F^{(0)} = \mathrm{MLP}_{F}\!\left( \tfrac{1}{2}\left(S_e + S_a\right) \right).
\label{eq:initial_fusion}
\end{equation}

This creates an initial aggregated feature by simply averaging the shared candidates ($S_e$ and $S_a$), which is then passed through a Multi-Layer Perceptron (MLP) that applies a non-linear transformation to form a compact latent fusion space. The resulting $F^{(0)}$ becomes the starting point for the proposed bi-directional feedback mechanism and acts as a shared communication channel between the two individual streams. It is then iteratively updated based on the feedback provided by $P_e$ and $P_a$.

\subsubsection{Bi-Directional Feedback Fusion}

Instead of relying on simple concatenation, our approach iteratively refines a global fused state through bi-directional feedback. This process unfolds in two steps: (i) the global state modulates each stream using context-aware scaling and shifting, and (ii) the streams feed updated information back to the global state. After $R$ rounds, the fused representation captures both long-term environmental context and short-term action-driven variations.

This iterative refinement is crucial in scenarios where sudden action-driven changes (e.g., PM$_{2.5}$ spikes from cooking or cleaning) occur alongside slower environmental trends such as CO$_2$ accumulation. By continuously exchanging information, the model dynamically adjusts its interpretation of actions based on evolving environmental conditions, leading to more robust and context-aware predictions.

Our proposed bi-directional feedback module builds on the initial fused representation $F^{(0)}$ (defined in \ref{eq:initial_fusion}) and performs $R$ rounds of feedback between the global fusion state $F^{(r)}$ and the stream-specific private subspaces, $P_e^{(r)}$ and $P_a^{(r)}$. This mechanism allows the global representation to iteratively refine how it modulates each stream while simultaneously being updated by information fed back from them.

At each iteration $r \in \{1,\ldots,R\}$, $F^{(r-1)}$ produces a pair of context-aware modulation parameters for both streams:

\begin{align}
[\boldsymbol{\gamma}_e^{(r)}, \boldsymbol{\beta}_e^{(r)}]
  &= F^{(r-1)} W^{(e)}_{\text{mod}} + b^{(e)}_{\text{mod}}, \\
[\boldsymbol{\gamma}_a^{(r)}, \boldsymbol{\beta}_a^{(r)}]
  &= F^{(r-1)} W^{(a)}_{\text{mod}} + b^{(a)}_{\text{mod}}.
\end{align}

The coefficients $\boldsymbol{\gamma}$ and $\boldsymbol{\beta}$ act as \emph{context-aware modulation gates}, scaling and shifting each stream’s private representation according to the evolving global context:
\begin{align}
P_e^{(r)} &= \tanh(\boldsymbol{\gamma}_e^{(r)}) \odot P_e^{(r-1)} + \boldsymbol{\beta}_e^{(r)},\\
P_a^{(r)} &= \tanh(\boldsymbol{\gamma}_a^{(r)}) \odot P_a^{(r-1)} + \boldsymbol{\beta}_a^{(r)}.
\end{align}

The bounded $\tanh(\cdot)$ activation ensures numerical stability by keeping modulation within $[-1,1]$. Through this context-aware process, the model determines how strongly each stream should contribute under different conditions—emphasizing rapid action-driven variations during PM$_{2.5}$ spikes or smoothing longer environmental trends in CO$_2$ accumulation.

The updated private representations are then fed back to refine the global state:

\begin{equation}
F^{(r)} = \mathrm{MLP}_{F}\!\left(
   F^{(r-1)}
   + P_e^{(r)} W_e
   + P_a^{(r)} W_a
   \right).
   \label{eq:feedback-update}
\end{equation}

This additive feedback ensures that $F$ evolves gradually, integrating slow-changing environmental cues with rapidly fluctuating action-induced information. After $R$ rounds, the final $F^{(R)}$ serves as a balanced, context-aware representation encoding the interplay between human activities and environmental dynamics.

\subsubsection{Dual Timescale Modules}

Once the bi-directional feedback is achieved, the final fused representation $F^{(R)}$ encodes contextual information from both the environmental and action streams. To translate this fused state into temporal pollutant forecasts, the model employs two complementary dual timescale modules, each specialised for a distinct temporal granularity.

\begin{itemize}
  \item \textbf{Long-term module (captures coarse-grained dynamics)}: Operates on the fused representation $F^{(R)}$ to capture slow-varying, low-frequency environmental patterns such as CO$_2$ accumulation, ventilation cycles, and background drifts. The corresponding recurrent process is given by,
  
  $Z_{\text{long}} = \mathrm{GRU}_{\text{long}}(F^{(R)})$.

  \item \textbf{Short-term module (captures fine-grained dynamics)} Processes the concatenation of the final private subspaces $[P_e^{(R)} \Vert P_a^{(R)}]$ to capture rapid, high-frequency variations driven by human activities, for example, cooking or cleaning. The short-term dynamics are modelled as,
  
  $Z_{\text{short}} =
  \mathrm{GRU}_{\text{short}}([P_e^{(R)} \Vert
  P_a^{(R)}])$.
\end{itemize}

The final fused temporal representation is obtained through concatenating the two temporal embeddings,

\begin{equation}
Z = [\,Z_{\text{long}} \Vert Z_{\text{short}}\,],
\end{equation}

which integrates both long-term and short-term dependencies. This dual timescale formulation enables the model to account for both gradual contextual evolution and transient activity-induced variations, thereby improving the fidelity of CO$_2$ and PM$_{2.5}$ forecasting.

\subsubsection{Prediction Head}

Once the temporal representation obtained from the dual time scale module, the next step is to map this embedding to the pollutant space, 

\begin{equation}
\hat{Y} = Z W_h + b_h
  \in \mathbb{R}^{L \times K},
\label{eq:pred-linear}
\end{equation}

where $W_h \in \mathbb{R}^{d_z \times K}$ and $b_h \in \mathbb{R}^{K}$ are learnable parameters, and $K=2$ corresponds to the CO$_2$ and PM$_{2.5}$ targets.

As the model is trained using a lookback window of length $L$ and a prediction horizon of length $P$, only the final $P$ timesteps of the output sequence are retained as the forecast,

\begin{equation}
\hat{Y}_{\text{pred}}
  = \hat{Y}_{L-P+1:L, :}
  \in R^{P \times K}.
\label{eq:pred-slice}
\end{equation}

This slicing operation extracts the forecasted segment corresponding to future pollutant values beyond the input horizon. In effect, the linear layer serves as a temporal readout head, converting latent sequence representations into physically interpretable pollutant concentrations for CO$_2$ and PM$_{2.5}$.

\subsection{Training Objectives and Loss Formulation}

The model is trained to minimize the discrepancy between the predicted pollutant sequence $\hat{{Y}}_{\text{pred}} \in \mathbb{R}^{P\times K}$ and the ground-truth future values $Y \in R^{P\times K}$, where $K=2$ for CO$_2$ and PM$_{2.5}$. The overall learning objective combines a primary prediction loss with auxiliary regularization and uncertainty-based components.

Our primary prediction loss is formulated as a combination of mean squared error (MSE) \cite{mse_loss} and the uncertainty aware loss modelled through the Gaussian negative log-likelihood (NLL) calculations. During the experiments we tested on  employing two types of uncertainty; Homoscedastic and Heteroscedastic. The uncertainty-based loss calculation can be explained as below.

\subsubsection{Uncertainty-Aware Loss Formulation}

Let the model produce a predictive mean vector $\boldsymbol{\mu}_t \in \mathbb{R}^{K}$ for each future time step $t$ and each target pollutant (i.e., CO$_2$ and PM$_{2.5}$), where $K$ is the number of prediction variables. In the uncertainty-aware setting, the model additionally predicts a log-variance term $\log \boldsymbol{\sigma}_t \in \mathbb{R}^{K}$, giving rise to a Gaussian predictive distribution,

\begin{equation}
    \hat{\mathbf{y}}_t \sim 
    \mathcal{N}\!\left(\boldsymbol{\mu}_t,\; 
    \mathrm{diag}(\boldsymbol{\sigma}_t^2)\right).
\end{equation}

To incorporate predictive uncertainty, we optimise the Gaussian negative log-likelihood: 

\begin{equation}
\label{eq:nll}
\mathcal{L}_{\mathrm{NLL}}
=
\frac{1}{T}\sum_{t=1}^{T}
\left[
    \frac{\left(\mathbf{y}_t - \boldsymbol{\mu}_t\right)^2}
         {2\,\boldsymbol{\sigma}_t^{2}}
    \;+\;
    \log \boldsymbol{\sigma}_t
    \;+\;
    \frac{1}{2}\log(2\pi)
\right].
\end{equation}

This loss penalises both \emph{over-confident} predictions (small $\sigma_t$ with large errors) through the quadratic term, and
\emph{under-confident} predictions (excessively large $\sigma_t$) through the $\log \sigma_t$ term. Thus, the model jointly balance accuracy and calibration.

The two types of uncertainity is defined by the way how $\boldsymbol{\sigma}_t$ is parameterised within the NLL formulation in Eq.~\eqref{eq:nll}.

\textbf{A. Homoscedastic Uncertainty (Shared Variance)}
\label{sec:homo}
In this setting, each pollutant is assigned a single global variance parameter, constant across all time steps. It can be defined as,
\begin{equation}
    \sigma_{t,k} = \sigma_{k}, \qquad \forall t.
\end{equation}
Equivalently,
\begin{equation}
    \log \sigma_k = \theta_k, \qquad \theta_k \in \mathbb{R}.
\end{equation}
Substituting it into Eq.~\eqref{eq:nll} results in, 
\begin{equation}
\mathcal{L}_{\mathrm{Homo}}
=
\frac{1}{T}\sum_{t,k}
\left[
    \frac{(y_{t,k} - \mu_{t,k})^2}{2e^{2\theta_k}}
    + \theta_k
\right] + C.
\end{equation}

\textbf{B. Heteroscedastic Uncertainty (MLP-Based)}
Here the variance is sample dependent which is predicted using an MLP layer. This is defined as,
\begin{equation}
    \log \boldsymbol{\sigma}_t = 
    f_{\mathrm{MLP}}({x}_t).
\end{equation}

This is then substituted into the Eq.~\eqref{eq:nll} as, 

\begin{equation}
\mathcal{L}_{\mathrm{Hetero}}
=
\frac{1}{T}\sum_{t,k}
\left[
    \frac{(y_{t,k} - \mu_{t,k})^2}{2\sigma_{t,k}^{2}}
    + \log \sigma_{t,k}
\right] + C.
\end{equation}

\subsubsection{Regularisation Terms}
Additionally, we define the regularisation terms in order to encourage the meaningful disentanglement between global and stream-specific representations as below.

Firstly, alignment between the global feature representations (${S}_e$, ${S}_a$) is essential because if the global representations are too different, the fusion will fail. By using cosine similarity, we ensure they point in similar directions

\begin{equation}
    \mathcal{R}_{\text{align}} = \mathbb{E}[\,1 - \cos({S}_e, {S}_a)\,],
\end{equation}

Secondly, the correlation between the global features (${F}^{(R)}$) and the private features (${P}_e$, ${P}_a$) needs to be penalized because the private features are required to be independent of the global features. This independence is encouraged by minimizing the covariance as follows:

\begin{equation}
    \mathcal{R}_{\text{ind}} = \|\mathrm{Cov}({F}^{(R)}, {P}_e)\|_F^2 + \|\mathrm{Cov}({F}^{(R)}, {P}_a)\|_F^2,
\end{equation}

Finally, it is important to ensure that the private features from different streams (${P}_e$, ${P}_a$) are diverse and contribute non-redundant information. This is encouraged through:

\begin{equation}
    \mathcal{R}_{\text{div}} = \mathbb{E}[\cos^2({P}_e,{P}_a)].
\end{equation}

These terms maintain a balance between shared (global) and private (stream-specific) components, ensuring that each stream contributes non-redundant information to the fusion process, thereby complementing the formulation of the proposed framework.

\subsubsection{6) Total Objective.}
The final loss ($\mathcal{L}_{\star}$) combines two complementary objectives,

\begin{equation}
    \mathcal{L}_{\star} =  \mathcal{L}_{\mathrm{MSE}} +\mathcal{L}_{\mathrm{NLL}}.
\end{equation}

The overall training objective integrates this prediction loss with regularization terms:

\begin{equation}
\mathcal{J}
 = \mathcal{L}_{\star}
 + \lambda_{\text{align}}\mathcal{R}_{\text{align}}
 + \lambda_{\text{ind}}\mathcal{R}_{\text{ind}}
 + \lambda_{\text{div}}\mathcal{R}_{\text{div}},
\label{eq:total-loss}
\end{equation}

This formulation supports flexible training strategies, such as warm-starting with MSE for stable convergence, followed by fine-tuning with NLL to incorporate uncertainty. Such a design reflects the physical and behavioral characteristics of indoor air-quality dynamics: deterministic trends are captured by MSE, while NLL accounts for stochastic variations and provides confidence estimates, critical for decision-support systems in ventilation control and health risk assessment.

\section{Experiments and Results}

In this section, we provide the details of the experimental evaluation conducted to assess the effectiveness, robustness, and design choices of the proposed dual‑stream bi-directional feedback fusion framework. We first describe the dataset characteristics, preprocessing steps, and evaluation metrics used to benchmark forecasting performance for both CO$_2$ and PM$_{2.5}$. We then present a comprehensive set of experiments examining forecasting accuracy across varying temporal horizons, followed by ablation studies isolating the impact of activity information, fusion strategies, temporal modelling choices, dual‑timescale modules, loss formulations, and regularisation components. Together, these analyses offer a systematic understanding of how each architectural element contributes to overall performance and provide empirical evidence for the advantages of integrating behavioural context with environmental sensing in IAQ prediction.

\subsection{Experimental Setup}
\subsubsection{Dataset and Preprocessing}

We use the DALTON indoor air quality dataset \cite{dalton_dataset}, the only large-scale dataset that combines pollutant measurements with detailed human activity information. DALTON provides long-term, multi-site recordings from 30 indoor environments, including homes, apartments, canteens, classrooms, and laboratories-collected over six months across summer and winter seasons. For this study, we focus on indoor sensor readings (temperature, humidity, CO$_2$, and PM$_{2.5}$) and real-time human activity annotations captured via a speech-to-text mobile application operated by occupants. These annotations enable modelling of activity-driven pollution dynamics, while the sensor data provide environmental context. DALTON also reflects real-world challenges such as missing data and operational variability, making it an ideal benchmark for evaluating action-aware forecasting models that integrate human behaviour with indoor pollutant trends.

\subsubsection{Evaluation Metrics}

To quantitatively assess the predictive performance of the proposed model for indoor CO$_{2}$ and PM$_{2.5}$ forecasting, we employ three commonly used regression metrics: Mean Squared Error (MSE), Root Mean Squared Error (RMSE), and the coefficient of determination (R²). These are calculated as follow, 

\begin{equation}
\mathrm{MSE} = \frac{1}{N} \sum_{i=1}^{N} \left( y_i - \hat{y}_i \right)^2,
\label{eq:mse}
\end{equation}

\begin{equation}
\mathrm{RMSE} = \sqrt{ \frac{1}{N} \sum_{i=1}^{N} \left( y_i - \hat{y}_i \right)^2 },
\label{eq:rmse}
\end{equation}

\begin{equation}
R^2 = 1 - 
\frac{\sum_{i=1}^{N} \left( y_i - \hat{y}_i \right)^2}
{\sum_{i=1}^{N} \left( y_i - \bar{y} \right)^2},
\label{eq:r2}
\end{equation}

where $y_i$ $\hat{y}_i$, $N$ and $\bar{y}$ denote the ground-truth values, the predicted values, total number of samples and the mean of the observed ground-truth values, respectively. We adopt MSE (Eq. ~\ref{eq:mse}), RMSE (Eq. ~\ref{eq:rmse}), and $R^2$ (Eq. ~\ref{eq:r2}) as complementary evaluation metrics, as each captures a distinct aspect of predictive performance. MSE places greater weight on larger prediction errors because of its squaring term, making it particularly responsive to periods when pollutant concentrations are elevated. This characteristic is especially valuable in indoor air quality forecasting, where health risks are often linked to sustained exposure to high PM$_{2.5}$ levels, and where elevated CO$_2$ concentrations serve as indicators of inadequate ventilation and a heightened likelihood of airborne disease transmission. Consequently, underestimating prolonged high‑concentration episodes can result in misleading exposure assessments and delayed implementation of necessary mitigation measures. RMSE retains the same physical units as the target variables (ppm for CO$_2$ and $\mu$g/m$^3$ for PM$_{2.5}$), enabling intuitive interpretation of average prediction errors under dynamic indoor conditions. In contrast, $R^2$ offers a relative measure of model performance by indicating the proportion of variance in pollutant concentrations that the model is able to explain. This makes it particularly valuable for comparing predictive accuracy across different monitoring sites and pollutant types.

\subsection{Experimental Results}
\subsection{Experiments under varying forecasting horizons}

In order to evaluate the proposed model’s ability to capture past measurements and forecast future pollutant levels, we experimented with varying lookback and prediction time horizons. The results are presented in Table \ref{tab:time_horizons}.

We observed that the best CO${2}$ prediction performance can be achieved through longer prediction horizons combined with longer lookback windows. Specifically, the highest performance was observed when the lookback window was $L{l}=120$ and the prediction horizon was $L_{p}=60$. For shorter horizons (e.g., $L_{p}=5, 10, 15$), RMSE and MAE are higher, and $R^2$ is slightly lower (0.89–0.93) compared to the longer setting. Interestingly, increasing $L_p$ generally improves $R^2$ for CO$_2$, especially when paired with longer lookback periods, although RMSE can spike sharply at intermediate horizons (e.g., $L_p = 30$ shows RMSE  148). This indicates compounding uncertainty over mid-range forecasts.

On the other hand, the best PM$_{2.5}$ performance was achieved at $L{l}=48$, $L_{p}=5$, with the highest $R^2$ of 0.88. Longer horizons significantly degrade PM$_{2.5}$ prediction accuracy, where RMSE jumps to 52–60 and $R^2$ drops to 0.53–0.68. Based on these results, we believe PM$_{2.5}$ is more sensitive to prediction horizon than CO$_{2}$.
Overall, a setting with a shorter prediction horizon of $L_{p}=15$ and a moderate lookback window of $L_{l}=48$ can be considered a configuration that works well for predicting the levels of both pollutants, CO$_2$ and PM$_{2.5}$.

\begin{table}[htbp!]
\centering
\caption{Experimental results across different forecasting horizons. Here, $L_{l}$ and $L_{p}$ denote the lookback and prediction windows, respectively.}
\small
\begin{tabularx}{\columnwidth}{l c ccc ccc}
\toprule
\multirow{2}{*}{$L_{l}$} & \multirow{2}{*}{$L_{p}$} & \multicolumn{3}{c|}{CO$_2$} & \multicolumn{3}{c}{PM$_{2.5}$} \\
\cmidrule(lr){3-5}\cmidrule(lr){6-8}
 &  & RMSE$\downarrow$ & MAE$\downarrow$ & R$^2$$\uparrow$ & RMSE$\downarrow$ & MAE$\downarrow$ & R$^2$$\uparrow$ \\
\midrule
 24& 5&  99.75 & 68.43 & 0.89 & 39.42 & 29.40 & 0.80 \\
36 &10 & 78.14& 40.43& 0.94& 32.63 &24.23 &0.86  \\
48 &15 & 81.47& 53.15&	0.93& \textbf{30.43} &\textbf{24.06} &\textbf{0.88}  \\
60 &15&  76.09& 50.97& 0.94 &35.42 &30.42 &0.84  \\
 60 & 30& 148.11 &82.01 &0.79 &52.32 &44.01 &0.65  \\
 90& 30& 148.11 &82.01 &0.79 &52.32 &44.01 &0.65 \\
 120 & 30&  128.59 &73.06 &0.84 &49.53 &42.19 &0.68  \\
 120& 60 & \textbf{68.90} &\textbf{41.85} &\textbf{0.96} &60.89 &52.42 &0.53  \\

\bottomrule
\end{tabularx}
\label{tab:time_horizons}
\end{table}


In an indoor setting CO$_{2}$ indoors is primarily driven by human occupancy and ventilation patterns. People exhale CO$_{2}$ at a fairly constant rate. Ventilation systems often follow predictable schedules (e.g., HVAC cycles). These factors create stable and periodic trends (e.g., peaks during working hours, drops at night). We believe the longer lookback windows ($L_{l}$ = 120) capture these cycles, and longer prediction horizons ($L_{p}$ = 60) remain accurate because occupancy and ventilation changes happens gradually. However, the behaviour of the PM2.5 indoors is highly event-driven. For example, the events such as cooking, cleaning and opening a window can cause sudden spikes. However, these events are short-lived and irregular. This has been the reason why shorter horizons ($L_{p}$= 15) with a moderate lookback window ($L_{l}$ = 48) work best as they capture recent activity without overextending into unpredictable future events.

\subsection{Impact of dual-stream information and the fusion strategy}
\label{sec:acts_and_fusions}

We conducted ablation experiments to determine the importance of integrating activity information and the proposed fusion mechanism. Here, we consider the setting with $L_{p}=15$, $L_{l}=48$, which provides overall strong prediction performance. As shown in Table \ref{tab:act_and_fusion}, the model that uses only environmental sensor readings (i.e., environmental stream $X_e$ only) has the worst performance compared to our proposed method with iterative feedback ($R=3$), where the R$^2$ gaps are 0.12 and 0.44 for CO$_2$ and PM$_{2.5}$ respectively.
Direct concatenation of the two streams ($X_e + X_a$) fails to improve performance because it does not allow modality-specific feature learning, resulting in poor integration of activity cues. This approach further drops R$^2$ by 0.04 and 0.07 from the single-stream setting for CO$_2$ and PM$_{2.5}$ respectively. In contrast, modeling the streams separately through individual encoders before concatenation significantly improves prediction performance, reducing RMSE by 21.24 for CO$_2$ and 5.68 for PM$_{2.5}$ compared to direct concatenation.
Through introducing bi-directional fusion without feedback instead of simple two-stream concatenation, we observed a major improvement, particularly for PM$_{2.5}$, where R$^2$ increased by 0.36 and RMSE dropped by 24.19 compared to the single-stream setting. For CO$_2$, RMSE decreased from 134.16 to 123.37 (an 8\% improvement).
We observed further improvements in predicting both pollutant levels through introducing feedback iterations ($R=3$). This refinement step reduced RMSE for CO$_2$ from 123.37 to 81.47 (a 39.3\% drop) and for PM$_{2.5}$ from 39.10 to 30.43 (a 22.1\% drop), while boosting R$^2$ to 0.93 and 0.88 respectively. These results highlight the progressive benefit of each design choice and signify the importance of the proposed bi-directional feedback-based fusion module. Feedback iterations allow the model to iteratively refine cross-stream interactions and capture temporal dependencies more effectively.
Overall, these findings suggest that activity-aware fusion is critical for pollutants influenced by human behavior (e.g., PM$_{2.5}$), while CO$_2$ benefits moderately from advanced fusion strategies.

\begin{table*}[htbp!]
\centering
\caption{Impact of dual-stream information and fusion strategy on IAQ forecasting performance.}
\small
\begin{tabularx}{0.65\textwidth}{l|ccc|ccc}
\toprule
\multirow{2}{*}{Model} & \multicolumn{3}{c|}{CO$_2$} & \multicolumn{3}{c}{PM$_{2.5}$} \\
\cmidrule(lr){2-4}\cmidrule(lr){5-7}
 & RMSE$\downarrow$ & MAE$\downarrow$ & R$^2$$\uparrow$ & RMSE$\downarrow$ & MAE$\downarrow$ & R$^2$$\uparrow$ \\
\midrule
$X_e$ only & 134.16& 71.27& 0.81& 65.60& 43.75& 0.44 \\
$X_e + X_a$ (direct concat) & 147.06 & 86.38 & 0.77 & 70.04 & 50.29 & 0.37 \\
$X_e + X_a$ (2-streams; concat) & 125.82 & 62.59 & 0.83 & 59.92 &39.63 & 0.54 \\
Ours (no feedback, $R{=}0$) & 123.37 & 	65.30 & 0.84 & 39.10 & 19.56 & 0.80 \\
\textbf{Ours (feedback, $R{=}3$)} & \textbf{81.47} &\textbf{ 53.15} & \textbf{0.93} & \textbf{30.43} & \textbf{24.06} &\textbf{ 0.88} \\
\bottomrule
\end{tabularx}
\label{tab:act_and_fusion}
\end{table*}

\subsection{Impact of different temporal modelling strategies}

In our proposed pollutant prediction framework, the temporal component within each individual stream encoder plays a major role in learning temporal patterns before the fusion stage. To identify the most effective temporal modeling strategy for this purpose, we tested multiple state-of-the-art temporal models and reported their impact on overall forecasting performance in Table \ref{tab:temporal_models}.

\begin{itemize}
    \item iTransformer \cite{liu2023itransformer}: A Transformer-based architecture designed for time-series forecasting. It leverages global self-attention to capture long-range dependencies across time steps, making it effective for periodic patterns such as CO$_2$ accumulation. However, its reliance on global attention can limit responsiveness to localized short-term fluctuations, which are common in PM$_{2.5}$ dynamics.
    
    \item Multi-Stage TCN \cite{farha2019ms}: A Temporal Convolutional Network variant that uses stacked dilated convolutions to model multi-scale temporal patterns. This approach efficiently captures short-term and medium-term dependencies with lower computational cost compared to Transformers. Nevertheless, its receptive field is limited, reducing its ability to model very long-term periodic trends.
    
    \item Patch-TST \cite{Yuqietal-2023-PatchTST}: A Transformer variant that processes time-series data in patches, similar to Vision Transformers for images. This patching strategy improves efficiency for long sequences and performs well for periodic signals like CO$_2$. However, fixed patch sizes can miss fine-grained short-term variations, making it less effective for irregular, event-driven PM$_{2.5}$ spikes.
\end{itemize}

As shown in Table \ref{tab:temporal_models}, incorporating advanced temporal models significantly aids the framework in achieving higher prediction accuracy. Among the tested approaches, our STE encoder provides the strongest contribution to overall performance, enabling the fused model to achieve RMSE of 81.47 and R$^2$ of 0.93 for CO$_2$, and RMSE of 30.43 with R$^2$ of 0.88 for PM$_{2.5}$. Compared to using Patch-TST within the stream encoder, this translates to an 18\% reduction in RMSE for CO$_2$ and a 13\% reduction for PM$_{2.5}$, along with improved R$^2$ scores.
Interestingly, MAE for PM$_{2.5}$ is slightly higher when using STE compared to Patch-TST (24.06 vs. 14.02), suggesting that Patch-TST handles average levels well but struggles with extreme spikes. In contrast, STE’s ability to combine efficient long-range memory with adaptive attention mechanisms helps capture both periodic CO$_2$ trends and sudden PM$_{2.5}$ fluctuations.
These findings highlight that the choice of temporal modelling strategy within stream encoders is critical for enabling the fusion module to leverage rich temporal representations. Models that balance long-term dependencies with adaptive responsiveness provide the most benefit for pollutants with mixed dynamics (periodic vs. event-driven).

\begin{table*}[htbp!]
\centering
\caption{Impact of different temporal modelling strategies on IAQ forecasting performance.}
\small
\begin{tabularx}{0.65\textwidth}{l|ccc|ccc}
\toprule
\multirow{2}{*}{Model} & \multicolumn{3}{c|}{CO$_2$} & \multicolumn{3}{c}{PM$_{2.5}$} \\
\cmidrule(lr){2-4}\cmidrule(lr){5-7}
 & RMSE$\downarrow$ & MAE$\downarrow$ & R$^2$$\uparrow$ & RMSE$\downarrow$ & MAE$\downarrow$ & R$^2$$\uparrow$ \\
\midrule
iTransformer & 128.23& 84.68& 0.82& 49.96& 23.52& 0.68 \\
Multi-Stage TCN (3 stages) & 120.68 & 73.02 & 0.84 & 41.16 & 33.04 & 0.78 \\
patch-TST & 100.03& 61.24& 0.89 &35.04 &14.02 &0.84 \\
\textbf{STE(Ours)} & \textbf{81.47}&\textbf{ 53.15}& \textbf{0.93}& \textbf{30.43}& \textbf{24.06}&\textbf{ 0.88} \\

\bottomrule
\end{tabularx}
\label{tab:temporal_models}
\end{table*}

\subsection{Impact of the bi-directional feedback fusion}


The fusion component plays a major role in our proposed two-stream framework. As an initial step, we experimented with varying fusion mechanisms, including direct concatenation, two-stream concatenation (as explained in Section \ref{sec:acts_and_fusions}), and the proposed bi-directional feedback fusion module with and without iterative feedback. Here, we compare the impact of different feedback iterations ($R$).
As shown in Table \ref{tab:fusion_ablation}, introducing feedback significantly improves performance compared to static fusion. For example, moving from direct concatenation to feedback with $R=3$ reduces RMSE for CO$_2$ from 147.06 to 81.47 (a 44.6\% improvement) and for PM$_{2.5}$ from 70.04 to 30.43 (a 56.6\% improvement), while R$^2$ increases from 0.77 to 0.93 and 0.37 to 0.88 respectively.
Experiments suggest that even a single feedback iteration ($R=1$) boosts accuracy significantly (CO$_2$ R$^2$ = 0.91), while the best performance is achieved at $R=3$. Beyond $R=3$, performance starts to degrade, likely due to over-refinement or noise amplification, which can destabilize predictions.
We believe the optimal feedback setting ($R=3$) strikes a balance between capturing temporal dependencies and avoiding overfitting or instability caused by excessive iterations. These findings highlight the importance of iterative refinement in fusion design, particularly for pollutants with mixed dynamics, where both long-term trends and short-term fluctuations must be modelled effectively.

\begin{table*}[!t]
\centering
\caption{Impact of bi-directional feedback fusion on IAQ forecasting performance.}
\begin{tabular}{l|ccc|ccc}
\toprule
\multirow{2}{*}{Fusion Variant} & \multicolumn{3}{c|}{CO$_2$} & \multicolumn{3}{c}{PM$_{2.5}$} \\
\cmidrule(lr){2-4}\cmidrule(lr){5-7}
 & RMSE$\downarrow$ & MAE$\downarrow$ & R$^2$$\uparrow$ & RMSE$\downarrow$ & MAE$\downarrow$ & R$^2$$\uparrow$ \\
\midrule
Direct Concat & 147.06 & 86.38 & 0.77 & 70.04 & 50.29 & 0.37  \\
Two-Streams Concat & 125.82 & 62.59 & 0.83 & 59.92 &39.63 & 0.54 \\
No feedback ($R{=}0$) & 123.37 & 65.30 & 0.84 & 39.10 & 19.56 & 0.80  \\
Feedback $R{=}1$ & 92.42 &47.40 &0.91 &35.04 &14.02 &0.84 \\
Feedback $R{=}3$ &  \textbf{81.47} & \textbf{53.15} &\textbf{ 0.93} & \textbf{30.43} & \textbf{24.06} & \textbf{0.88 } \\
Feedback $R{=}5$ & 85.74& 50.93 &0.92 &31.29 &17.23 &0.87  \\
Feedback $R{=}7$  &  96.34& 57.30& 0.90& 33.71& 25.18& 0.85  \\

\bottomrule
\end{tabular}
\label{tab:fusion_ablation}
\end{table*}

\subsection{Impact of Dual-Timescales} 

Once the bi-directional feedback fusion is performed, the dual-timescale module is responsible for capturing both long-term and short-term temporal patterns, enabling the model to handle pollutants with mixed dynamics. Table \ref{tab:dual_timescale} compares the prediction performance of the dual-timescale module against simplified counterparts that focus on capturing either short-term or long-term dependencies.
When comparing the results, the short-term-only module effectively captures rapid dynamics, resulting in better PM$_{2.5}$ predictions but noisier CO$_2$ forecasts. In contrast, the long-term-only module learns slow accumulative patterns of CO$_2$ but misses transient activity-driven spikes of PM$_{2.5}$. The dual-timescale module combines these complementary strengths, achieving the best accuracy and generalization. Specifically, RMSE improves by 17.8\% for CO$_2$ (from 99.15 to 81.47) and by 20.7\% for PM$_{2.5}$ (from 38.39 to 30.43), while R$^2$ rises to 0.93 and 0.88 respectively.
These findings confirm the necessity of hierarchical temporal modelling for multi-pollutant indoor air-quality forecasting, as it enables simultaneous handling of gradual CO$_2$ accumulation trends and sudden PM$_{2.5}$ spikes caused by short-lived events.

\begin{table*}[!t]
\centering
\caption{Impact of dual-timescale temporal modules on forecasting performance.}
\begin{tabular}{l|ccc|ccc}
\toprule
\multirow{2}{*}{Temporal Variant} & \multicolumn{3}{c|}{CO$_2$} & \multicolumn{3}{c}{PM$_{2.5}$} \\
\cmidrule(lr){2-4}\cmidrule(lr){5-7}
 & RMSE$\downarrow$ & MAE$\downarrow$ & R$^2$$\uparrow$ & RMSE$\downarrow$ & MAE$\downarrow$ & R$^2$$\uparrow$ \\
\midrule
Short-term only &  100.39 &74.22 &0.90  &36.65	&28.23	&0.83  \\
Long-term only &  99.15	&71.23	&0.90 &38.39 &29.20	&0.81  \\
With dual-timescale &  \textbf{81.47} & \textbf{53.15} &\textbf{ 0.93} & \textbf{30.43} & \textbf{24.06} & \textbf{0.88}  \\
\bottomrule
\end{tabular}
\label{tab:dual_timescale}
\end{table*}

\subsection{Impact of Loss Formulation}

The loss formulation plays a critical role in shaping the prediction behaviour of our framework. Table \ref{tab:loss_ablation} compares three strategies: using $\mathcal{L}_{\mathrm{MSE}}$ alone, combining MSE with negative log-likelihood (NLL) under homoscedastic uncertainty, and combining MSE with NLL under heteroscedastic uncertainty.
Adding NLL introduces probabilistic modelling, enabling the network to capture prediction uncertainty. While homoscedastic NLL assumes constant uncertainty across all predictions, heteroscedastic NLL adapts uncertainty dynamically based on input conditions, which is particularly useful for pollutants with irregular patterns.
As shown in the table, the heteroscedastic formulation achieves the best overall performance. Compared to MSE-only, RMSE improves by 13\% for CO$_2$ (93.61 → 81.47) and by 36\% for PM$_{2.5}$ (47.88 → 30.43), while R$^2$ rises from 0.91 to 0.93 for CO$_2$ and from 0.70 to 0.88 for PM$_{2.5}$. This demonstrates that uncertainty-aware loss functions significantly enhance prediction accuracy, especially for PM$_{2.5}$, which exhibits high variability due to short-lived events.
Interestingly, MAE for CO$_2$ increases slightly when adding NLL (43.05 → 53.15), suggesting that the model prioritizes reducing large errors (RMSE) over small ones when uncertainty is modelled. These findings highlight that loss formulation is not just a training detail but a key design choice for improving robustness and generalization in multi-pollutant forecasting.

\begin{table*}[!t]
\centering
\caption{Impact of loss formulation on IAQ forecasting performance.}
\begin{tabular}{l|ccc|ccc}
\toprule
\multirow{2}{*}{Loss} & \multicolumn{3}{c|}{CO$_2$} & \multicolumn{3}{c}{PM$_{2.5}$} \\
\cmidrule(lr){2-4}\cmidrule(lr){5-7}
 & RMSE$\downarrow$ & MAE$\downarrow$ & R$^2$$\uparrow$ & RMSE$\downarrow$ & MAE$\downarrow$ & R$^2$$\uparrow$ \\
\midrule
$\mathcal{L}_{\mathrm{MSE}}$ only & 93.61 &43.05 &0.91 & 47.88 & 35.69 & 0.70 \\
$\mathcal{L}_{\mathrm{MSE}} +\mathcal{L}_{\mathrm{NLL}}$ (Homo)  & 87.83 & 51.53 & 0.92 & 45.13 & 37.39 & 0.74\\
$\mathcal{L}_{\mathrm{MSE}} +\mathcal{L}_{\mathrm{NLL}}$ (Hetero) &  \textbf{81.47} & \textbf{53.15} &\textbf{ 0.93} & \textbf{30.43} & \textbf{24.06} & \textbf{0.88 } \\

\bottomrule
\end{tabular}
\label{tab:loss_ablation}
\end{table*}

\subsection{Impact of the regularisation terms}

Regularization plays a critical role in stabilizing training and improving generalization in our multi-stream framework. Table \ref{tab:regular_ablation} compares the impact of progressively adding three regularization terms (discussed in Sec).

As shown in the table, adding $\mathcal{R}{\text{align}}$ to the base loss ($\mathcal{L}_{\star}$) significantly improves PM$_{2.5}$ predictions (RMSE drops from 52.10 to 37.75, a 27.5\% improvement), while CO$_2$ remains relatively stable. Introducing $\mathcal{R}{\text{ind}}$ further enhances CO$_2$ performance (RMSE reduces from 101.09 to 83.12, an 18\% improvement). Finally, incorporating $\mathcal{R}{\text{div}}$ yields the best overall results, reducing RMSE to 81.47 for CO$_2$ and 30.43 for PM$_{2.5}$, while R$^2$ rises to 0.93 and 0.88 respectively.

These findings highlight that regularization is not just a safeguard against overfitting but a mechanism for enforcing complementary learning across streams. The combination of alignment, independence, and diversity constraints ensures that the fusion module leverages both shared and unique information effectively, leading to robust multi-pollutant forecasting.

\begin{table*}[!t]
\centering
\caption{Impact of regularisation terms on IAQ forecasting performance}
\begin{tabular}{l|ccc|ccc}
\toprule
\multirow{2}{*}{Regularisation} & \multicolumn{3}{c|}{CO$_2$} & \multicolumn{3}{c}{PM$_{2.5}$} \\
\cmidrule(lr){2-4}\cmidrule(lr){5-7}
 & RMSE$\downarrow$ & MAE$\downarrow$ & R$^2$$\uparrow$ & RMSE$\downarrow$ & MAE$\downarrow$ & R$^2$$\uparrow$ \\
\midrule
$\mathcal{L}_{\star}$ only & 105.11	&55.49	&0.89	&52.10	&30.19	&0.65  \\
$\mathcal{L}_{\star}$ + $\mathcal{R}_{\text{align}}$  & 101.09	&66.23	&0.89	&37.75	&30.59	&0.82 \\
$\mathcal{L}_{\star}$ + $\mathcal{R}_{\text{align}}+ \mathcal{R}_{\text{ind}}$ & 83.12
 &38.12 &0.92 &	36.82&	26.35&	0.82 \\
$\mathcal{L}_{\star}$ + $\mathcal{R}_{\text{align}}+ \mathcal{R}_{\text{ind}}+ \mathcal{R}_{\text{div}}$ &  \textbf{81.47} & \textbf{53.15} &\textbf{ 0.93} & \textbf{30.43} & \textbf{24.06} & \textbf{0.88 } \\

\bottomrule
\end{tabular}
\label{tab:regular_ablation}
\end{table*}

\subsection{Qualitative Results}

\begin{figure*}[h]
        \centering
        	\includegraphics[width=0.8\textwidth]{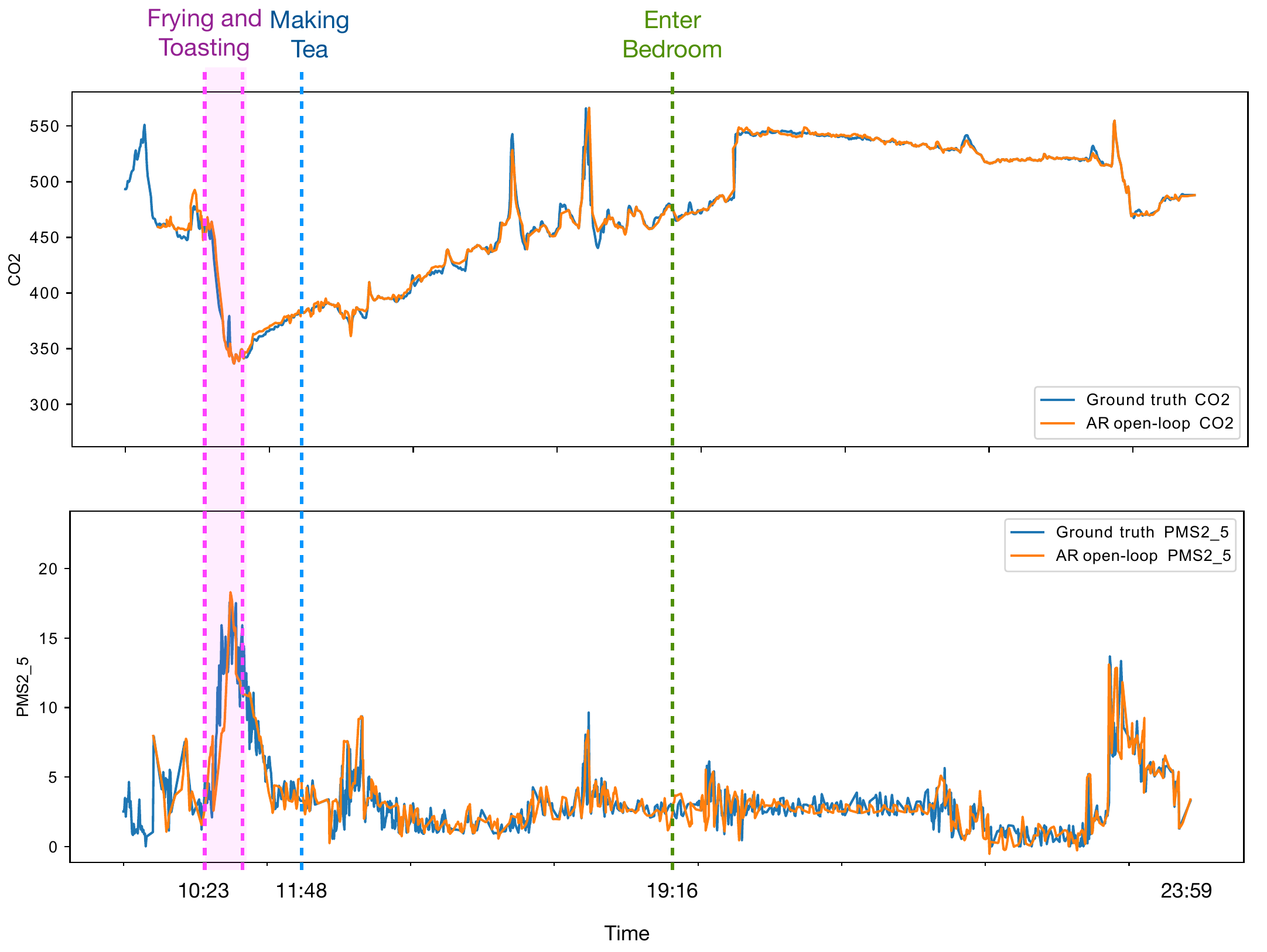}
	\caption{}
	\label{fig:plots1}
\end{figure*}

Figure \ref{fig:plots1} presents the temporal evolution of indoor CO$_2$ concentration and PM$_{2.5}$ levels over a full day, together with corresponding predictions generated by our proposed model. Several household activities such as frying, toasting, tea preparation, and bedroom occupancy are annotated to highlight their influence on pollutant dynamics.

In this study, we use an open‑loop autoregressive model, which is a simple method that predicts the next value of a signal based only on its past values. Unlike closed‑loop methods, it does not update or correct itself using new measurements as it goes. We tested this approach because it is practical for real‑world deployments where sensors may be low‑cost, data may drop out, or continuous calibration is not possible. This makes open‑loop models useful in situations where a lightweight and reliable prediction method is needed without constant feedback from sensors.

The measured CO$_{2}$ concentrations display clear, event‑driven changes, with increases of approximately 200 ppm during these events. Noticeable increases appear during activities such as cooking and spending time in the bedroom, which is consistent with what we know about indoor CO$_{2}$ sources, mainly human breathing and combustion from cooking. The proposed model predictions follows the general rise and fall of CO$_{2}$ reasonably well, capturing the main trends over time. However, it struggles during sharp peaks: the model tends to underestimate these higher values and shows a slight delay compared to the actual measurements. The predicted signal also appears smoother, which suggests the model filters out quick changes. This is expected, as open‑loop models do not correct themselves using new measurements and therefore accumulate error over time.

For PM$_{2.5}$, the measurements show more variability, with values ranging from 0 to about 20 µg/m³, and the concentrations are clearly influenced by specific events. The strongest increase occurs during frying, which matches existing evidence that frying produces large amounts of fine particles. Toasting and making tea also create smaller increases, while time spent in the bedroom shows almost no PM$_{2.5}$ change, reflecting the absence of particle‑generating activities. Unlike CO$_{2}$, PM$_{2.5}$ levels change quickly and unpredictably, making them harder to model. Although the AR open‑loop predictions capture the general timing of events, they fail to match the actual peak heights. The frying peak, in particular, is significantly underestimated, and the model outputs a smoother, slightly delayed version of what is measured.

Overall, the AR open‑loop model performs well for broad, slow‑moving trends and correctly identifies when major events happen. It also produces stable predictions that do not drift over time. 

\section{Conclusion}
This work presented a bi-directional feedback fusion framework for indoor air‑quality forecasting that explicitly integrates environmental sensing with action-aware embeddings to model the joint influence of environmental conditions and human behaviour on pollutant dynamics. Motivated by the limitations of conventional forecasting approaches, which primarily exploit historical sensor trajectories while overlooking latent behavioural drivers, we formulated IAQ prediction as a dual-stream temporal modelling task with iterative cross‑modal refinement.

Extensive experiments on the DALTON dataset demonstrate that occupant activity information provides a critical complementary signal, particularly for pollutants with irregular, event-driven patterns such as PM$_{2.5}$. Across all evaluation settings, models relying solely on environmental streams showed the weakest performance, whereas introducing activity representations and the proposed bi‑directional feedback module yielded substantial gains. The iterative fusion process was especially effective, with three feedback rounds consistently producing the best results before additional iterations introduced instability. These findings validate the importance of progressive cross‑stream alignment when modelling mixed pollutant dynamics driven by both long-term environmental cycles and short-lived human actions.

Our analysis further highlights the necessity of multi-scale temporal reasoning. The dual-timescale module provided robust improvements over single‑scale alternatives by capturing both the gradual, predictable evolution of CO$_{2}$ and the rapid fluctuations characteristic of PM$_{2.5}$. Likewise, uncertainty‑aware loss formulations, particularly heteroscedastic NLL, proved essential for improving spike responsiveness and enhancing reliability under volatile conditions. Together, these components enable the model not only to achieve state‑of‑the‑art accuracy but also to generate interpretable uncertainty estimates valuable for real‑world decision support.

Overall, the proposed framework advances the state of IAQ forecasting by demonstrating that meaningful integration of behavioural context fundamentally reshapes model expressiveness and robustness. By combining dual-stream temporal encoders, bi‑directional feedback fusion, hierarchical timescale modelling, and uncertainty‑aware optimisation, the approach provides a unified foundation for predicting pollutants with diverse dynamics. Future work will explore closed-loop forecasting, broader behavioural signal sources (e.g., wearables, vision, or passive sensing), and real-time deployment within smart ventilation and health-monitoring systems.

\section*{Acknowledgments} 
This paper was supported by the ARC Training Centre for Advanced Building Systems Against Airborne Infection Transmission (THRIVE) (IC220100012), and ARC Laureate Fellowship (FL220100082).

\bibliographystyle{IEEEtran}
\bibliography{references.bib}

\newpage

 





\end{document}